\documentclass[10pt,twocolumn,letterpaper]{article}

\usepackage[arxiv]{cvpr}     

\usepackage{graphicx}
\usepackage{amsmath,amssymb,amsthm,mathabx}
\usepackage{booktabs}
\usepackage{algorithmic}
\usepackage[linesnumbered,ruled,vlined]{algorithm2e}
\usepackage{acronym}
\usepackage{enumitem}
\usepackage[colorlinks,urlcolor=blue,bookmarks=false,linkcolor=blue,citecolor=blue,urlcolor=blue,pagebackref=true]{hyperref}
\usepackage{balance}
\usepackage{xspace}
\usepackage{setspace}
\usepackage[skip=3pt,font=small]{caption}
\usepackage[capitalise]{cleveref}
\usepackage{tabularx,colortbl,multirow,array,makecell}
\usepackage{overpic,wrapfig}
\usepackage{xr}

\makeatletter
\newcommand*{\addFileDependency}[1]{
  \typeout{(#1)}
  \@addtofilelist{#1}
  \IfFileExists{#1}{}{\typeout{No file #1.}}
}
\makeatother

\makeatletter
\DeclareRobustCommand\onedot{\futurelet\@let@token\@onedot}
\def\@onedot{\ifx\@let@token.\else.\null\fi\xspace}
\def\eg{\emph{e.g}\onedot} 

\def\ie{\emph{i.e}\onedot}

\def\etal{\emph{et al}\onedot}
 
\makeatother

\graphicspath{{figures/}}

\crefname{algocf}{Alg.}{Algs.}
\Crefname{algocf}{Algorithm}{Algorithm}

\acrodef{hint}[\textsc{Hint}]{\underline{H}andwritten arithmetic with \underline{INT}egers}
\acrodef{ans}[ANS]{Arithmetic Neural-Symbolic}
\acrodef{nn}[NN]{Neural Network}

\usepackage{calc}
\newlength\myheight
\newlength\mydepth
\settototalheight\myheight{Xygp}
\newlength\savewidth\newcommand\shline{\noalign{\global\savewidth\arrayrulewidth\global\arrayrulewidth 1pt}\hline\noalign{\global\arrayrulewidth\savewidth}}

\def\cvprPaperID{4330} 
\def\confName{CVPR}
\def\confYear{2022}

\definecolor{mborange}{rgb}{1.0,0.5,0.0}

\begin{document}
\title{Towards a Unified Foundation Model: \\ Jointly Pre-Training  Transformers on Unpaired Images and Text}  

\author{Qing Li$^{1,2}$
\and
Boqing Gong$^{1}$
\and
Yin Cui$^{1}$
\and
Dan Kondratyuk$^{1}$
\and
Xianzhi Du$^{1}$
\and
Ming-Hsuan Yang$^{1}$
\and
Matthew Brown$^{1}$
\and
\\
$^{1}$Google Research~~~~~$^{2}$University of California, Los Angeles  
}

\maketitle

\begin{abstract}
In this paper, we explore the possibility of building a unified foundation model that can be adapted to both vision-only and text-only tasks. Starting from BERT and ViT, we design a unified transformer consisting of modality-specific tokenizers, a shared transformer encoder, and task-specific output heads. To efficiently pre-train the proposed model jointly on unpaired images and text, we propose two novel techniques: (i) We employ the separately-trained BERT and ViT models as teachers and apply knowledge distillation to provide additional, accurate supervision signals for the joint training; (ii) We propose a novel gradient masking strategy to balance the parameter updates from the image and text pre-training losses. We evaluate the jointly pre-trained transformer by fine-tuning it on image classification tasks and natural language understanding tasks, respectively. The experiments show that the resultant unified foundation transformer works surprisingly well on both the vision-only and text-only tasks, and the proposed knowledge distillation and gradient masking strategy can effectively lift the performance to approach the level of separately-trained models.

\end{abstract}

\section{Introduction}
In recent years, deep learning models for different domains, particularly computer vision and natural language processing, have gradually converged to very similar paradigms in terms of both model architectures and learning methods: transformers~\cite{vaswani2017attention} learnt in a pre-training-fine-tuning fashion, which are named \textit{foundation models} \cite{bommasani2021opportunities}. In the NLP community, previous works (\eg, \cite{devlin2019bert,liu2019roberta,brown2020language,raffel2019exploring}) have demonstrated that transformers, which are pre-trained on large text corpora, achieve state-of-the-art performance when fine-tuned on a wide range of downstream textual tasks. In the visual domain, transformer-based models (\eg, \cite{parmar2018image,dosovitskiy2020image,touvron2021training,carion2020end,liu2021swin}) have recently achieved promising results on image classification, object detection, and panoptic segmentation. These transformers are usually pre-trained on a large image datasets and then fine-tuned on the specific tasks.

However, despite the achievements of transformers on various domains, we still have to maintain a large foundation model for each domain or modality~\cite{bommasani2021opportunities}. This limits the impact of such models and is contrary to the long-pursued goal of artificial general intelligence, which aims to simultaneously handle multiple modalities in an efficient way. After witnessing the paradigm shift of deep learning to transformer-based foundation models, various questions naturally arise: Can we build a unified transformer to solve tasks across different modalities? If not, is it possible to pre-train a single transformer, which then can be fine-tuned to tasks of different modalities? 

In this paper, we take a step toward positively answering the above questions by exploring approaches to building a single pre-trained transformer for visual and textual modalities, which can be further fine-tuned to different downstream tasks. Starting from BERT~\cite{devlin2019bert} and ViT~\cite{dosovitskiy2020image}, we propose a unified foundation transformer, named ``ViT-BERT'', consisting of three modules: (i) Modality-specific tokenizers: each modality has its own tokenizer to process the specific form of inputs; (ii) Shared transformer encoder: all modalities and tasks share a single transformer as the main body of the model; (iii) Task-specific heads: each task has a lightweight classifier to predict the output for the task at hand. The design philosophy is to have \textit{minimal} modality-specific and task-specific parameters and as many parameters and computation as possible in the shared transformer encoder. Therefore, we adopt single-layer patch projection and embeddings for the modality-specific tokenizers and a two-layer MLP for each task-specific head. We pre-train the proposed unified transformer on \textit{unpaired} images and text (\eg, Wikipedia). The model then serves as a unified foundation model that transfers well to both vision-only and text-only tasks. 

To efficiently pre-train the unified foundation transformer, we propose two novel techniques. First, to provide rich, accurate supervision signals for the joint pre-training, we exploit the separately-trained BERT and ViT as teacher models and apply knowledge distillation to train the proposed model. Second, to reconcile the potentially conflicting gradients from different tasks, we design a novel gradient masking strategy to balance the learning signals from the text and image pre-training tasks. More concretely, we create a mask to select the most important set of parameters for the text pre-training based on the magnitude of its gradients, and leave the remaining parameters to be updated by the image pre-training. The proposed gradient masking strategy is gradually applied during the training process until a desired sparsity of the mask is reached. 

We evaluate the representation power of the jointly pre-trained ViT-BERT model by fine-tuning it on vision-only and text-only tasks. More specifically, we fine-tune the model on image classification tasks, such as CIFAR-10/100 and ImageNet, to test how well it can transfer to vision-only tasks taking images as inputs. We also fine-tune it on the GLUE benchmark~\cite{wang2018glue} to test its capability on natural language understanding. The experimental results show that jointly pre-training on images and text works surprisingly well and does not cause a significant performance drop on the downstream tasks, compared with the separately-trained BERT and ViT. The proposed knowledge distillation and gradient masking strategy can further improve the performance to approach the level of separately-trained models, validating their effectiveness on pre-training the unified foundation transformer. 
\section{Related Work}
Our work is related to transformer-based foundation models and multi-task learning, which we discuss below.

\subsection{Transformer-based Foundation Models}
The Transformer, which is based on a multi-headed self-attention mechanism, was first invented in the natural language processing (NLP) community for machine translation tasks~\cite{vaswani2017attention}. Because of its flexibility and scalability, transformers have been rapidly adopted for large-scale language pre-training. Many works from the NLP community, such as BERT~\cite{devlin2019bert}, RoBERTa~\cite{liu2019roberta}, ALBERT~\cite{lan2019albert}, GPT~\cite{radford2018improving,radford2019language,brown2020language}, and T5~\cite{raffel2019exploring}, have demonstrated that transformer-based models pre-trained on large text corpora learn generic representations for natural language that can be transferred to a wide range of downstream tasks via fine-tuning and boost the performance by a large margin.

Recently, transformer-based models have been introduced to  computer vision. Image Transformer~\cite{parmar2018image}, iGPT~\cite{chen2020generative}, ViT~\cite{dosovitskiy2020image}, DETR~\cite{carion2020end}, and Swin Transformer~\cite{liu2021swin} and other recent works have applied transformers to various vision tasks and achieved competitive results on image classification, object detection, and panoptic segmentation.

Transformer-based models are also widely used for joint vision-and-language reasoning tasks. VisualBERT~\cite{li2019visualbert}, VL-BERT~\cite{su2019vl}, ViLBERT~\cite{lu2019vilbert}, LXMERT~\cite{tan2019lxmert}, and UNITER~\cite{chen2020uniter} have shown that being pre-trained on large-scale image-text pairs, transformer-based models can outperform previous models by a large margin in a wide range of multi-modal tasks, such as visual question answering, image captioning, and image-text retrieval. Note that these models require an object detection backbone (\eg, Faster-RCNN~\cite{ren2015faster}) to extract region features for images, which are then fed into transformers.

These transformer-based models are sometimes called \textit{foundation models}~\cite{bommasani2021opportunities}, which are learned by a pre-training and fine-tuning process. The pre-training is usually performed on large generic datasets, such as Wikipedia, JFT-300M, or Conceptual Captions~\cite{sharma2018conceptual}, and then pre-trained models are fine-tuned on downstream tasks with relatively small specific datasets.

\subsection{Multi-task Learning}
Multi-task learning~\cite{caruana1997multitask,crawshaw2020multi}, which aims to develop models that can handle multiple tasks whilst sharing parameters and computation between them, has been a long-standing research problem in machine learning, natural language understanding~\cite{hashimoto2016joint,liu2019multi,clark2019bam}, computer vision~\cite{he2017mask,zamir2018taskonomy,strezoski2019many,zamir2020robust}, or multi-modal reasoning~\cite{kaiser2017one,pramanik2019omninet,lu202012,hu2021unit,akbari2021vatt}. For example, Mask-RCNN~\cite{he2017mask} can handle object detection, segmentation, and pose estimation in a single network. Liu \etal \cite{liu2019multi} proposed MT-DNN, a multi-task language understanding model built by sharing lower layers in a transformer while making the top layer task-specific.

While the majority of previous works on multi-task learning focus on only a single  modality, there are also several notable prior works on a unified model for multiple  modalities. To simultaneously handle multiple tasks across vision and language (\ie, image classification, machine translation, image captioning \textit{etc.}), Kaiser \etal \cite{kaiser2017one} presented a heterogeneous model consisting of convolutional layers to process images, and attention and mixture-of-experts layers to model natural language. Hu \etal \cite{hu2021unit} proposed UniT, a unified transformer-based encoder-decoder architecture that handles multiple tasks and modalities in a single model. Note that the UniT model only shares the decoder among different tasks, while the encoder is specialized for each modality. In particular, they adopt a pre-trained DeTR~\cite{carion2020end} as the visual encoder and a pre-trained BERT~\cite{devlin2019bert} as the text encoder. Akbari \etal \cite{akbari2021vatt} proposed VATT to process video, audio, and text via a single transformer encoder, which is trained by multi-modal contrastive learning and desires aligned data triplets across modalities.

There are two major differences between our paper and previous works on multi-task learning. First, we target a \textit{unified foundation} model that is adaptable to a wide range of tasks in different modalities and domains. Therefore, multi-task learning in our work occurs at the pre-training stage, instead of the task-specific training phase when most previous works perform multi-task learning for various  tasks directly. Second, we intentionally design a transformer-based model with \textit{minimal} modality-specific parameters. Specifically, the proposed ViT-BERT model has single-layer patch projection or embeddings for tokenizing images and text, which are much more lightweight than the separate encoders used in UniT~\cite{hu2021unit}.
\section{Method}
\begin{figure*}[t]
	\centering
	\includegraphics[width=\linewidth]{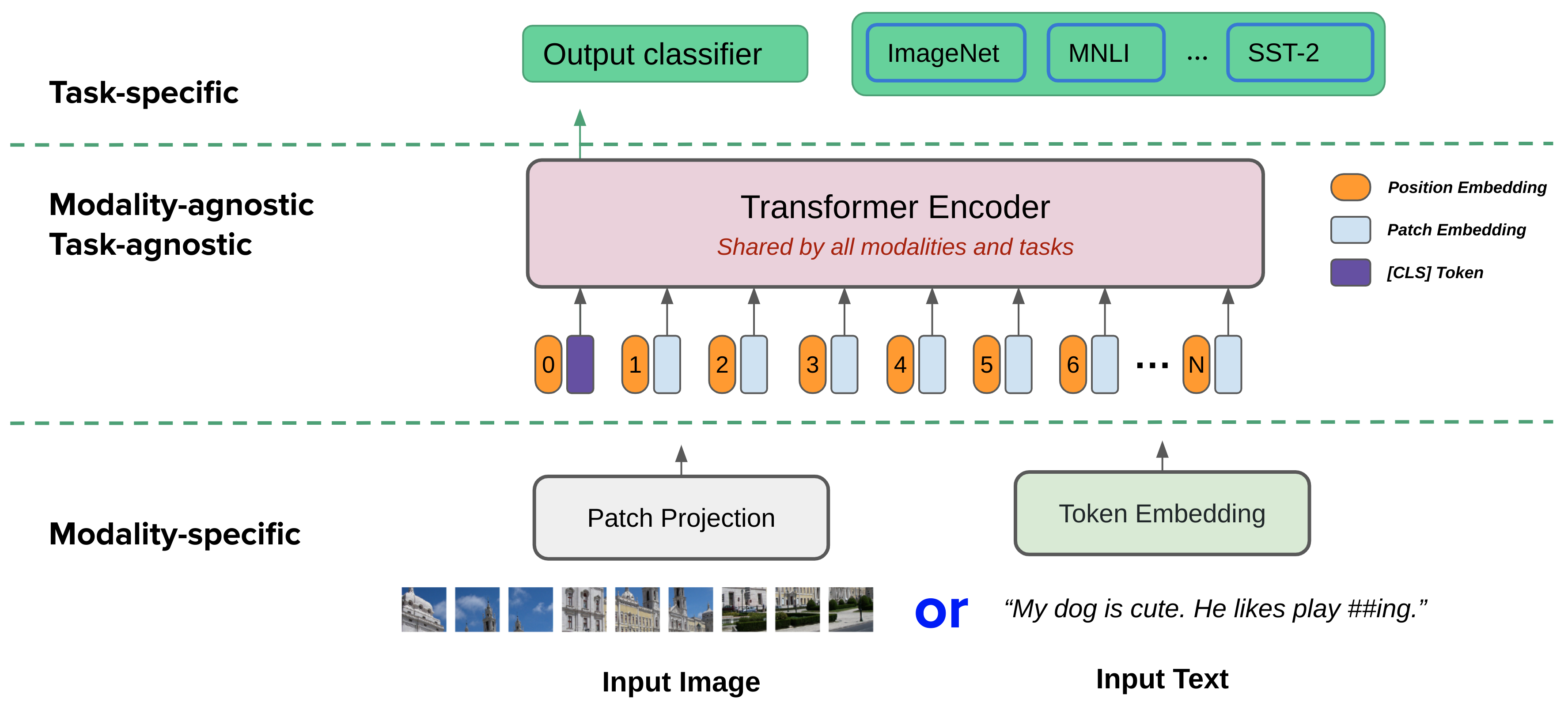}
	\caption{An illustration of the proposed ViT-BERT model. Our model consists of three modules: (i) modality-specific tokenizers, which can take in image inputs or text inputs; (ii) modality-agnostic task-agnostic transformer encoder, which is shared by all modalities and tasks; (iii) task-specific heads, which generate predictions for different tasks.}
	\label{fig:framework}
\end{figure*}


\subsection{Unified Transformer}
\subsubsection{Modality-specific Tokenizer}
We consider two input modalities: either images or text. Note that we do not use image-text pairs for pre-training, so the following formulation does not consider paired images and text as input. 

The vision-only tasks require perceiving images as input. Inspired by the Vision Transformer (ViT)~\cite{dosovitskiy2020image}, we tokenize an image $I \in \mathbb{R}^{H \times W \times C}$ by first slicing it into a sequence of patches $V \in \mathbb{R}^{N \times (P^2 \times C)}$, where $(P,P)$ is the patch resolution and $N = HW/P^2$ is the resulting number of patches. Then, the patches are each flattened and embedded by a linear projection $V \in \mathbb{R}^{(P^2 \times C) \times D}$, finally prepended with a class token $v_{cls}$ and added with a position embedding $V^{pos} \in \mathbb{R}^{N \times D}$. Formally, an image is processed into a sequence of patch embeddings as follows:
\begin{align}
I = [v_{cls}, v_1 V, \ldots, v_N V] + V^{pos}
\end{align}

In the text-only tasks, the input text (\eg, a single sentence or a pair of sentences) is tokenized in the same way as in BERT into a sequence of tokens as following: 
\begin{align}
S = [t_{cls}, t_1 T, \ldots, t_M T] + T^{pos} + T^{seg}
\end{align}
where $T$ is the word embedding matrix, $T^{pos}$ is the position embedding for text, and $T^{seg}$ is the segment embedding.

We do not add any extra embedding for modality types because it brings no improvement in our experiment.

\subsubsection{Shared Transformer Encoder}
The transformer encoder consists of stacked blocks including a multiheaded self-attention (MSA) layer and a MLP layer. The MLP contains two fully-connected layers with a GELU non-linearity. Layer normalization (LN) is applied before each MSA or MLP layer:
\begin{align}
    &z^0 = I \text{ or } S \nonumber \\
    &\hat{z}^l = \text{MSA}(\text{LN}(z^{l-1})) + z^{l-1},~&l=1, \ldots,L  \\
    &z^l = \text{MLP}(\text{LN}(\hat{z}^l)) + \hat{z}^l,~&l=1,\ldots,L
\end{align}
The final output of the transformer encoder is the token embedding $z^L$ at the last layer, which is used as input to different task-specific heads.

\subsubsection{Task-specific Heads}
A task-specific prediction head is applied over the final output of the transformer encoder for each task. All tasks that we address in this work, including both the pre-training tasks and the downstream tasks, can be cast into a classification problem. To predict the output classes, we apply a two-layer MLP classifier with GeLU activation and the hidden dimension is equal to the hidden size of the transformer encoder. More formally, we have:
\begin{align}
    &z = W_2 \cdot \text{GeLU}(W_1 \cdot z^L_0 + b_1) + b_2,\\
    &p = \text{Softmax}(z)
\end{align}
where $W_1, b_1$ are the weights and bias of the first layer, $W_2, b_2$ are the weights and bias of the second layer, and $p$ is the predicted probability distribution over the output classes. Note that $z^L_0$ is the embedding of the class token at the last layer of the transformer in most tasks, except for the masked language modelling, in which it is the embedding of the masked token at the last layer.

\subsection{Joint Pre-Training}
In this section we discuss the joint pre-training of the proposed model on images and text. The joint pre-training belongs to the regime of multi-task learning, which has long been believed to be very challenging due to optimization problems like conflicting gradients~\cite{yu2020gradient} or catastrophic forgetting~\cite{kirkpatrick2017overcoming}. Jointly training a neural network to simultaneously perform multiple tasks has typically required careful calibration of the individual tasks, to ensure that none of the task-specific losses dominates another. These problems are even more severe in our case because the pre-training is very noisy and usually requires millions of optimization steps to converge, especially for the pre-training on text. 

We address the above issues from two perspectives: (i) we use \textbf{knowledge distillation} to provide extra accurate supervision for the joint training; (ii) we design a \textbf{gradient masking} strategy to accommodate the potentially conflicting gradients from different tasks. These two techniques are described next in detail.

\subsubsection{Knowledge Distillation} \label{sec:kd}
In this section we discuss how knowledge distillation is used to improve the joint pre-training on text and images. We assume we have access to the original BERT and ViT models, which are separately pre-trained on the text or image modalities, as two teacher models for the proposed unified model. The question to be addressed here is how to conduct training by exploiting these two teachers.

Knowledge distillation minimizes the Kullback-Leibler (KL) divergence between the probability distributions of a teacher model and the student model. Let $z_t$ be the predicted logits of the teacher model, $z_s$ the predicted logits of the student model, and $y$ be the ground-truth label. The distillation objective is:
\begin{align} \label{eq:distill}
\mathcal{L} = (1 - \alpha) \mathcal{L}_\text{CE}(\psi(z_s), y) + \alpha \text{KL}(\psi(\frac{z_s}{\tau}), \psi(\frac{z_t}{\tau}))
\end{align}
where $\psi$ denotes the softmax function, $\alpha$ is the ratio balancing the cross-entropy loss ($\mathcal{L}_\text{CE}$) over the ground-truth label and the KL divergence, and $\tau$ is the temperature hyperparameter used in distillation.

Since we are training the student model simultaneously on text and images, each training batch includes both text and images. We then have two loss terms derived from \autoref{eq:distill}: $\mathcal{L}_{\text{img}}$ for the image pre-training task and $\mathcal{L}_{\text{txt}}$ for the text pre-training task. We can simply sum up these two losses and calculate the combined gradients to update our model or  combine the gradients from these two losses  in a way that avoids their conflicts, as described next.

\subsubsection{Gradient Masking}
As described in the previous section, we have two loss terms for the image pre-training and the text pre-training, respectively. Since the two optimize for different modalities and objectives, they could generate conflicting gradients for the joint training. Simply adding the two losses, which means totally ignoring the gradient conflict, may slow down the training process and render the model optimization sub-optimal~\cite{yu2020gradient}. 

Therefore, instead of directly adding these two losses, we explore new ways to accommodate the conflicting gradients by drawing inspiration from the neural network pruning literature~\cite{han2015learning,liu2018rethinking,frankle2018lottery,you2019drawing}. It is commonly believed that current large neural networks are highly over-parameterized, and techniques for eliminating unnecessary weights from neural networks (\ie, pruning) have been shown to be able to reduce the parameter counts of trained networks by even 90\% without compromising accuracy. 

Inspired by these results from neural network pruning, we propose a novel gradient masking strategy to reconcile the potentially conflicting gradients from the text pre-training and the image pre-training. The main idea is that we can keep a subset of most important gradients for the text pre-training and ignore other gradients to leave room for the image pre-training. More formally, let $\theta$ be the parameters in the shared transformer encoder, $\mathcal{L}_{\text{txt}}, \mathcal{L}_{\text{img}}$ be the losses of the text pre-training task and the image pre-training task, respectively. We combine the gradients from the text and image pre-training tasks by an adapted mask $M$. Formally, we have:
\begin{align}
    & G_{\text{txt}}  = \frac{ \partial{\mathcal{L}_{\text{txt}}} } {\partial{\theta}},  \hspace{2mm}
    G_{\text{img}}  = \frac{ \partial{\mathcal{L}_{\text{img}}} } {\partial{\theta}}  \nonumber \\
    & G_{\text{global}}  = M \odot G_{\text{txt}} + (1-M) \odot G_{\text{img}}
\end{align}
Intuitively, the mask $M$ is supposed to select the most important gradients for the text pre-training and leave the rest for the image pre-training, so we heuristically generate the mask $M$ based on the magnitude of $G_{\text{txt}}$. Inspired by the Iterative Magnitude Pruning algorithm~\cite{frankle2018lottery,liu2018rethinking} used in the network pruning literature, we design an iterative gradient masking procedure to gradually increase the sparsity of the mask, as detailed in \autoref{alg:gm}. The iterative gradient masking procedure brings negligible extra computation cost because the mask $M$ is only updated for a few times during the whole training process, \eg, if the final masking ratio ($\beta$) is set to 50\% and the masking ratio per iteration ($\delta$) is set to 10\%, the mask will be only updated for 5 times. Therefore, the gradient masking strategy does not slow down the training process.

\begin{algorithm}[h] 
    \small
    \caption{Iterative Gradient Masking}
    \label{alg:gm}
    \begin{algorithmic}[1] 
    \STATE \textbf{Input}: transformer encoder $\theta$, total training steps $T$, \\
    mask $M$, masking ratio $\beta$, mask updating interval $t$, \\
    masking ratio per iteration $\delta$.
    \STATE Initialize $\theta$ from the pre-trained BERT.
    \STATE Initialize the mask $M$ to all ones.
    \REPEAT
        \STATE Sample a batch and calculate the gradients $G_{\text{txt}}$ and $G_{\text{img}}$
        \STATE Prune the smallest $\delta$ of the non-zero elements of $M \odot |G_{\text{txt}}|$ and update $M$ accordingly.
        \STATE Train $\theta$ for $t$ steps using the gradient $G_{\text{global}} = M \odot G_{\text{txt}} + (1-M) \odot G_{\text{img}}$
    \UNTIL{the sparsity of $M$ reaches $\beta$: \ie, $\sum M / ||\theta|| = \beta$}
    \STATE Continue the training to the total training steps $T$.
    \end{algorithmic}
\end{algorithm}

\section{Experiments and Results}\label{sec:experiments}
\begin{table*}[]
\small
\centering
\begin{tabular}{l|cccccc|cccccc|c}
\shline
\multirow{2}{*}{\textbf{Pre-train}} & \multicolumn{6}{c|}{\textbf{Text-only Tasks}}                                                 & \multicolumn{6}{c|}{\textbf{Vision-only Tasks}}                                                    & \multicolumn{1}{l}{\multirow{2}{*}{\textbf{Avg.}}} \\
                                & \textbf{MNLI} & \textbf{QQP} & \textbf{QNLI} & \textbf{SST-2} & \textbf{RTE} & \textbf{Avg.} & \textbf{C10} & \textbf{C100} & \textbf{IN-1K} & \textbf{Flower} & \textbf{Pet} & \textbf{Avg.} & \multicolumn{1}{l}{}\\
\hline
Rand. Init. &43.4	&75.5	&49.4	&80.5	&51.6	&60.1 &68.0	&39.2	&19.5	&27.8	&9.1	&32.7 &46.4 \\
BERT                       & 84.4          & 90.6         & 91.0          & 92.5           & 81.9         & \textbf{88.1}          & 68.8         & 38.0          & 17.9              & 16.3            & 9.5          & 30.1          & 59.1                                               \\
ViT                        & 51.6          & 75.7         & 49.5          & 68.0           & 52.0         & 59.4          & 98.1         & 87.1          & 77.9              & 89.5            & 93.8         & \textbf{89.3}          & 74.3                                               \\
Joint                     & 78.4          & 88.4         & 87.4          & 80.5           & 60.3         & 79.0          & 97.5         & 84.8          & 74.2              & 88.5            & 93.0         & 87.6          & 83.3                                               \\ \hline
ViT-BERT                        & 81.8          & 88.9         & 89.7          & 90.4           & 64.6         & 83.1          & 98.3         & 84.8          & 78.3              & 89.7            & 93.7         & 89.0          & \textbf{86.0}                                               \\
~~~~~w/o KD                     & 81.9          & 89.6         & 90.0          & 89.2           & 58.8         & 81.9          & 97.4         & 82.3          & 75.4              & 88.2            & 92.2         & 87.1          & 84.5                                               \\
~~~~~w/o GM                     & 80.8          & 89.6         & 88.9          & 91.1           & 57.4         & 81.6          & 97.6         & 85.3          & 79.1              & 90.3            & 92.8         & 89.0          & 85.3                                               \\
\shline
\end{tabular}
\caption{The accuracy of fine-tuning on the downstream text-only tasks and vision-only tasks. For MNLI, we show accuracy averaged over the matched and mismatched sets. The results for BERT on text-only tasks and for ViT on vision-only tasks are cited from~\cite{iki2021effect} and \cite{dosovitskiy2020image}, respectively. Highlighted in bold is the highest accuracy of the average of text-only tasks, vision-only tasks, and all tasks across all models.}
\label{tab:all}
\end{table*}

\subsection{Tasks and Datasets} \label{sec:tasks}
\subsubsection{Image Modality}
We use the ILSVRC-2012 ImageNet dataset~\cite{deng2009imagenet} with 1K classes and 1.3M images as the image pre-training dataset. The image pre-training task is image classification with a resolution of $224 \times 224$  and the pre-trained model is fine-tuned on several image classification benchmarks: ImageNet-1K (IN-1K), CIFAR-10/100 (C10/100)~\cite{krizhevsky2009learning}, Oxford-IIIT Pets (Pet)~\cite{parkhi2012cats}, and Oxford Flowers-102 (Flower)~\cite{nilsback2008automated}.

Following~\cite{dosovitskiy2020image}, we fine-tune all models using SGD with a momentum of 0.9. All models are fine-tuned with cosine learning rate decay, a batch size of 512, no weight decay, and gradient clipping at global norm 1. The fine-tuning image resolution is $384 \times 384$ and the fine-tuning steps are 20,000 for ImageNet-1k, 10,000 for CIFAR-10/100, and 500 for Oxford-IIIT Pets and Oxford Flowers-102.

\subsubsection{Text Modality}
Following~\cite{devlin2019bert}, we use the concatenation of BooksCorpus (800M words)~\cite{zhu2015aligning} and English  Wikipedia (2,500M words) as the text pre-training corpus. For Wikipedia, we extract only the text passages and ignore lists, tables, and headers. The text pre-training tasks are masked language modelling and next sentence prediction. The pre-trained model is fine-tuned on five tasks (MNLI, QQP, QNLI, SST-2, and RTE) from the General Language Understanding Evaluation (GLUE) benchmark~\cite{wang2018glue}: Multi-Genre Natural Language Inference (MNLI), Quora Question Pairs (QQP), Question Natural Language Inference (QNLI), Stanford  Sentiment  Treebank (SST-2), and Recognizing  Textual  Entailment (RTE). 

All tasks in the text modality are fine-tuned using the AdamW optimizer~\cite{loshchilov2017decoupled}, a weight decay of 0.01, and a polynomial learning rate decay. All models are fine-tuned on each task for 3 epochs. We report the accuracy on the development sets of all tasks.

\subsection{Model Variants}
We name our full model as ViT-BERT, which is based on a 12-layer transformer with 768 hidden size and 3072 MLP size. For image inputs during pre-training, we adopt the 224x224 resolution with a fixed patch size of 16x16, resulting in a patch sequence of length 14x14. During fine-tuning, we feed images of the 384x384 resolution because it is often beneficial to fine-tune at higher resolution than pre-training. To take in images with higher resolution, we use the same patch size as pre-training and perform 2D interpolation of the pre-trained position embeddings to handle longer sequence than pre-training. For the textual input, we use WordPiece embeddings with a vocabulary size of 30,000. The position embeddings are learnt with a max sequence length of 512.

The ablated model without knowledge distillation is denoted as ``ViT-BERT-w/o KD'' and the one without gradient masking as ``ViT-BERT-w/o GM''. For baselines, we compare our model with the original BERT~\cite{devlin2019bert} and ViT~\cite{dosovitskiy2020image}. We also build a joint training baseline (``Joint'') by simply adding up the losses from the text pre-training objectives and the image pre-training objectives. A randomly-initialized baseline without pre-training (``Rand. Init.'') is provided for comparison; note that we still fine-tune it for each downstream task.

We fine-tune the above pre-trained models on both text-only and image-only tasks. Note that when the original BERT is fine-tuned on image-only tasks, only the transformer is initialized from the BERT while the patch projection and the position embeddings are randonly-initialized. It is similar for fine-tuning ViT on text-only tasks, in which the word embeddings and the position embeddings are randomly initialized.

\subsection{Pre-training Setting}
All models are pre-trained for 1M steps using the AdamW optimizer~\cite{loshchilov2017decoupled} with $\beta_1 = 0.9, \beta_2 = 0.999$ and the weight decay of 0.01. The learning rate is linearly warmed up for the first 10K steps to achieve a peak value of $5 \times 10^{-4}$ and then decayed in a polynomial schedule. We mix the two pre-training datasets within each batch, containing 4,096 images and 512 textual documents. We use a dropout probability of 0.1 on all layers and adopt the gelu activation~\cite{hendrycks2016gaussian}.

In the proposed ViT-BERT model, we set the distillation ratio ($\alpha$) to 1 and the temperature ($\tau$) to 1 in the knowledge distillation, and set the mask ratio ($\beta$) to 50\%, the mask update interval ($t$) to 10K, the mask ratio per iteration ($\delta$) to 10\% in the iterative gradient masking, if these hyperparameters are not mentioned explicitly.

\subsection{Results and Analyses}

\subsubsection{Fine-tuning on Downstream Tasks}

To examine the quality of the joint pre-training on images and text, we first compare our models by fine-tuning on the downstream tasks described in \autoref{sec:tasks}. \autoref{tab:all} shows the fine-tuning results on both the text-only tasks and the vision-only tasks. By inspecting the results in \autoref{tab:all}, we draw the following observations:
\begin{enumerate}[wide, labelindent=0pt]
    \item \textit{The proposed ViT-BERT model successfully learns a good representation for both the vision-only and the text-only downstream tasks.} \\
    As shown in \autoref{tab:all}, the proposed ViT-BERT model achieves very competitive performance on the vision-only tasks with an average accuracy of 89.0\%, compared to the original ViT model with an average accuracy of 89.3\%. Moreover, ViT-BERT obtains slightly higher accuracy than ViT on the CIFAR-10, ImageNet, and Oxford Flowers-102 datasets. For the text-only tasks, ViT-BERT obtains an average accuracy of 83.1\%, which is about five percent lower than that of the original BERT model (88.1\%). By inspecting the detail accuracy on each textual task, we found that the performance gap between ViT-BERT and BERT mainly comes from the RTE task. This is probably because RTE has much less data for fine-tuning, compared with the other four textual tasks.
    
    \item \textit{Cross-modal fine-tuning brings negative impact.}\\
    As we may expect, neither fine-tuning ViT on text-only tasks (Avg.: 59.4\%) nor fine-tuning BERT on vision-only tasks (Avg.: 30.1\%) works well. They even perform worse than the Rand. Init. baseline without any pre-training. This means that the modality gap between vision and language cannot easily be bridged by the fine-tuning. Performing cross-modal fine-tuning may even hurt the performance probably due to the mismatch of input forms and training objectives between pre-training and fine-tuning.
    
    \item \textit{The joint training baseline performs surprisingly well on both vision-only and text-only tasks.}\\
    The joint training baseline (``Joint'') achieves an average accuracy of 79.0\% on text-only tasks and an average accuracy of 87.6\% on vision-only tasks. Although still a large gap compared to BERT on text-only tasks and ViT on image-only tasks, the Joint baseline gains significant improvement compared to directly performing cross-modal fine-tuning. This observation suggests that the visual and textual modalities can share a representation space in the transformer.
    
    \item \textit{Both knowledge distillation and gradient masking benefit the joint training on images and text.}\\
    By inspecting the results of ViT-BERT and its ablated variants, we can see that using the knowledge distillation alone (``ViT-BERT w/o GM'') can bring a large performance gain across almost all tasks, compared with the naive joint training baseline. Particularly, knowledge distillation boosts the accuracy of SST-2 from 80.5\% to 91.1\% and the accuracy of IN-1K from 74.2\% to 79.1\%. These results demonstrate that the knowledge distillation can  improve the joint training on images and text by providing more accurate learning signals for both the text and image pre-training. \vspace{2pt}\\ 
    By comparing the performance of ViT-BERT w/o KD and Joint, we see that the gradient masking strategy can improve the performance on text-only tasks from an average accuracy of 79\% to 81.9\%, while the performance of vision-only tasks only drops slightly by 0.5\%. This demonstrates the advantage of gradient masking on balancing the potentially conflict gradients from the text pre-training objectives and the image pre-training objectives. \vspace{2pt} \\ 
    By combing knowledge distillation and gradient masking, the proposed ViT-BERT model can obtain an accuracy of 86\% averaged over all tasks, which is a 2.7 percent improvement over the joint training baseline (83.3\%).
\end{enumerate}

\subsubsection{Ablation Study}
In this section we conduct ablation studies to explore the impact of hyperparameters in the ViT-BERT model.

We first study how the distillation ratio and temperature influence the model performance. As shown in \autoref{tab:kd}, it consistently benefits both text-only and vision-only tasks to increase the distillation ratio, which means giving more weight to the KL loss term in \autoref{eq:distill}. Particularly, setting the distillation ratio to 0.7 obtains highest accuracy on text-only tasks, while a distillation ratio of 1 (\ie, only the KL loss term) works the best for the vision-only tasks. In terms of the distillation temperature, 1 is the best choice for both text-only and vision-only tasks and the model performs much worse when the distillation temperature exceeds 10. For simplicity, we set both the distillation ratio and temperature to 1 in the proposed ViT-BERT model.

\begin{table}[ht]
\centering
\begin{tabular}{cc|ccc}
\shline
\textbf{Ratio} & \textbf{Temp.} & \textbf{Text-only} & \textbf{Vision-only} & \textbf{Avg.} \\
\hline
0              & 1              & 79.0               & 87.6                 & 83.3          \\
0.3            & 1              & 78.9               & 87.9                 & 83.4          \\
0.5            & 1              & 79.4               & 88.4                 & 83.9          \\
0.7            & 1              & \textbf{81.7}      & 88.7                 & 85.2          \\
1              & 1              & 81.6               & \textbf{89.0}        & \textbf{85.3} \\
1              & 0              & 80.8               & 88.0                 & 84.4          \\
1              & 0.1            & 78.5               & 86.5                 & 82.5          \\
1              & 10             & 75.7               & 88.9                 & 82.3          \\
1              & 100            & 67.2               & 75.3                 & 71.3          \\
\shline
\end{tabular}
\caption{The impact of distillation ratio ($\alpha$) and temperature ($\tau$) in the knowledge distillation (\autoref{eq:distill}). Above experiments are conducted without gradient masking. Setting the distillation ratio to 0 denotes no distillation (\ie, the ``Joint'' baseline) and setting the distillation temperature to 0 means that we use hard distillation, \ie, the hard labels predicted by the teacher models are used as supervision to the trained model.}
\label{tab:kd}
\end{table}

\begin{figure}[t]
	\centering
	\includegraphics[width=\linewidth]{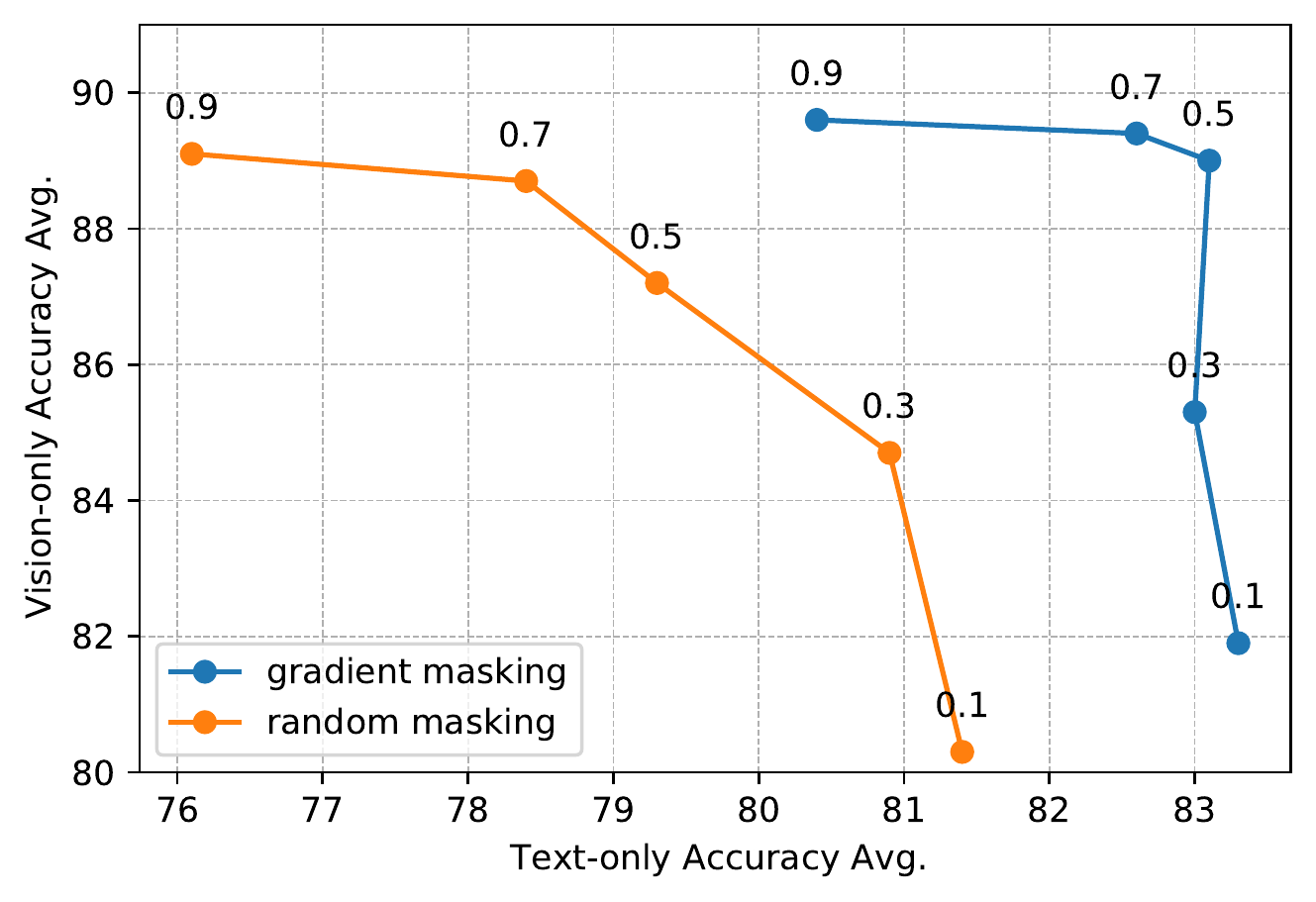}
	\caption{The impact of the mask ratio ($\beta$ in \autoref{alg:gm}) on the accuracy of downstream tasks. The numbers near the dots are the masking ratios used to train the models. ``random masking'' is a baseline using randomly-generated masks. Note that knowledge distillation is also used in these models.}
	\label{fig:gm}
\end{figure}

Another important hyperparamter in our model is the masking ratio used in the gradient masking strategy. \autoref{fig:gm} shows the accuracy of the ViT-BERT model with different masking ratios. For comparison, we also add a ``random masking'' baseline that randomly generates masks, instead of generating masks based on the gradient magnitude. From \autoref{fig:gm}, we can see that the models using gradient masking perform significantly better than those using random masking. This indicates that the masks in the ViT-BERT model, which are generated based on the gradient magnitude, indeed play a vital role in effectively combining the potentially conflict gradients from the text pre-training and the image pre-training. A masking ratio around 0.5 is a good compromise between the accuracy of text-only tasks and the accuracy of vision-only tasks, while increasing the mask ratio to 0.9 would drop the average accuracy of text-only tasks by about 3 percent and decreasing the ratio to 0.1 would lower the average accuracy of vision-only tasks by about 7 percent.

\subsection{Qualitative Study}
\begin{figure*}[t] \vspace{-5mm}
	\centering
	\includegraphics[width=\linewidth]{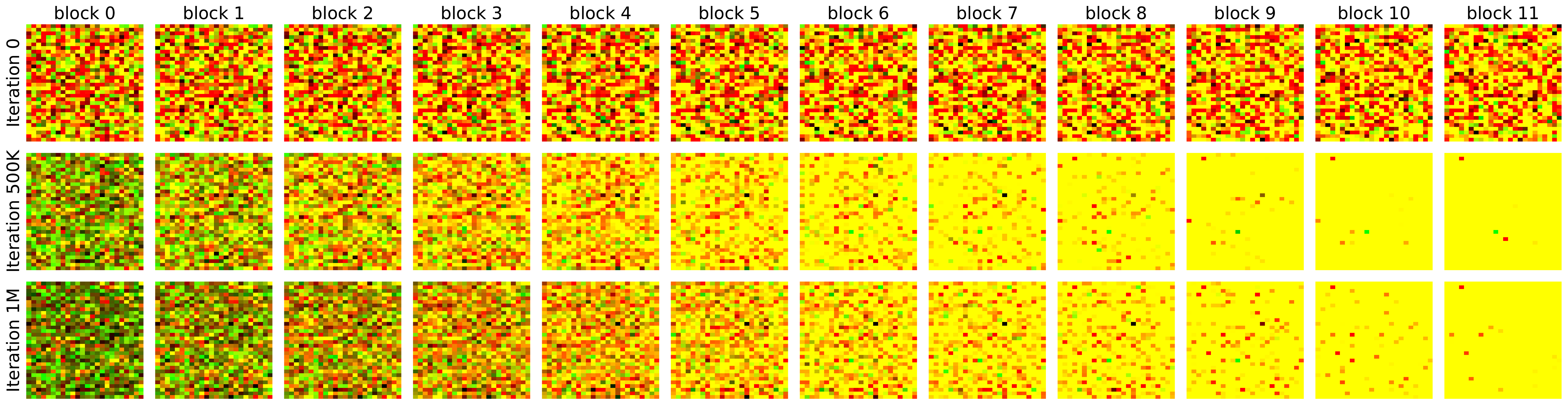}
	\caption{An illustration of the neuron activation of transformer blocks at different steps of training. Each sub-heatmap represents the activation output of a transformer block: the first channel (red) of the heatmap is the neuron activation maximized over 100 randomly-sampled images, the second channel (green) is the neuron activation maximized over 100 randomly-sampled text, and the third channel is set to all zeros. Therefore, \colorbox{red}{\color{red}e}{ }(red) implies the neuron has a high response to at least one image, \colorbox{green}{\color{green}e}{ }(green) implies the neuron has high a response to at least one text, \colorbox{yellow}{\color{yellow}e}{ }(yellow = red + green) implies the neuron has high responses to both an image and a text, and \colorbox{black}{\color{black}e}{ }(black) means that the neuron has no response to either images or text.}
	\label{fig:neuron_activation}
\end{figure*}

\begin{figure}[ht]
	\centering
	\includegraphics[width=\linewidth]{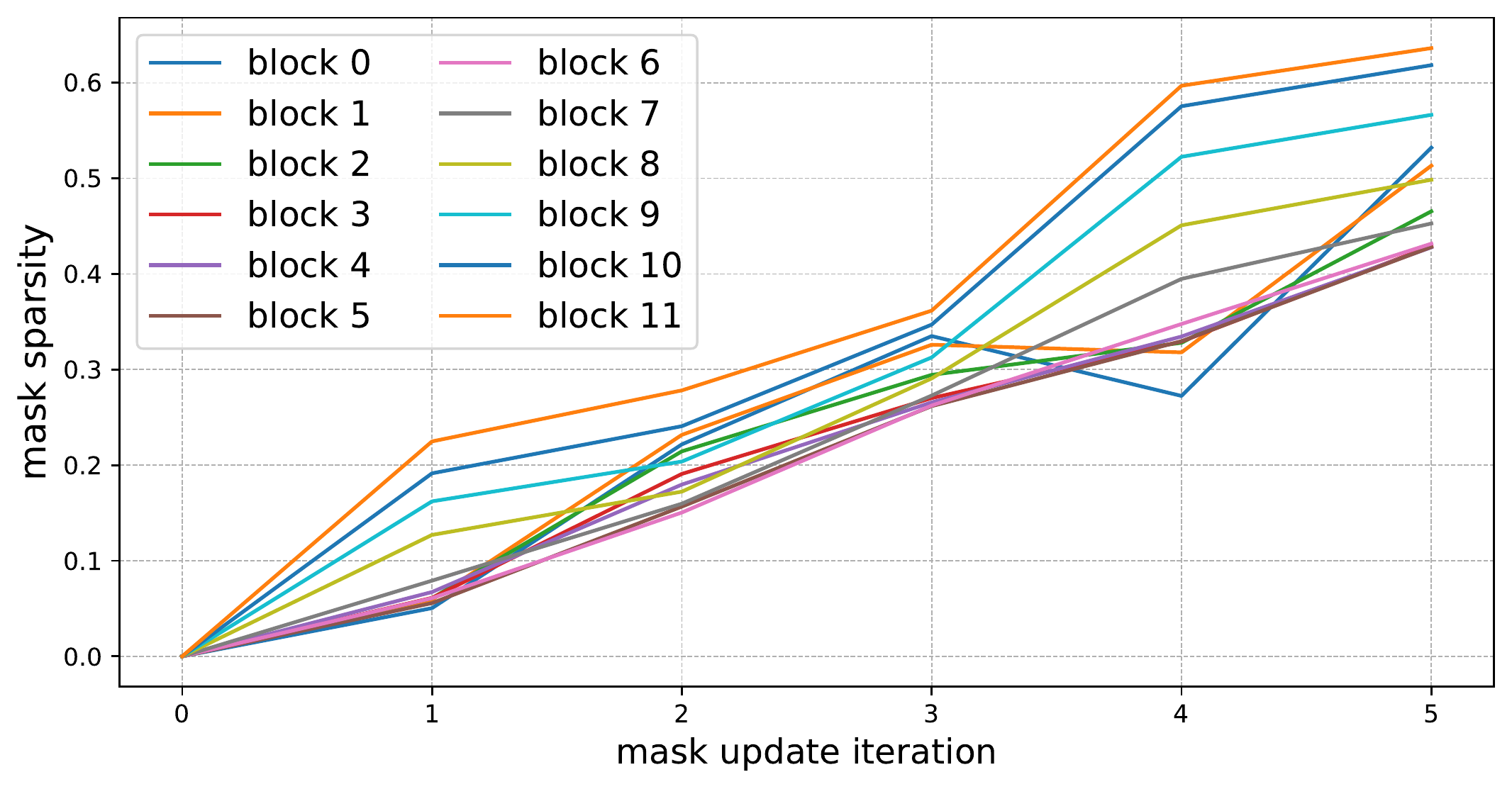}
	\caption{The sparsity of masks for all blocks in the transformer encoder at different iterations of mask updating.}
	\label{fig:mask_sparsity}
\end{figure}

\begin{figure}[ht]
	\centering
	\includegraphics[width=\linewidth]{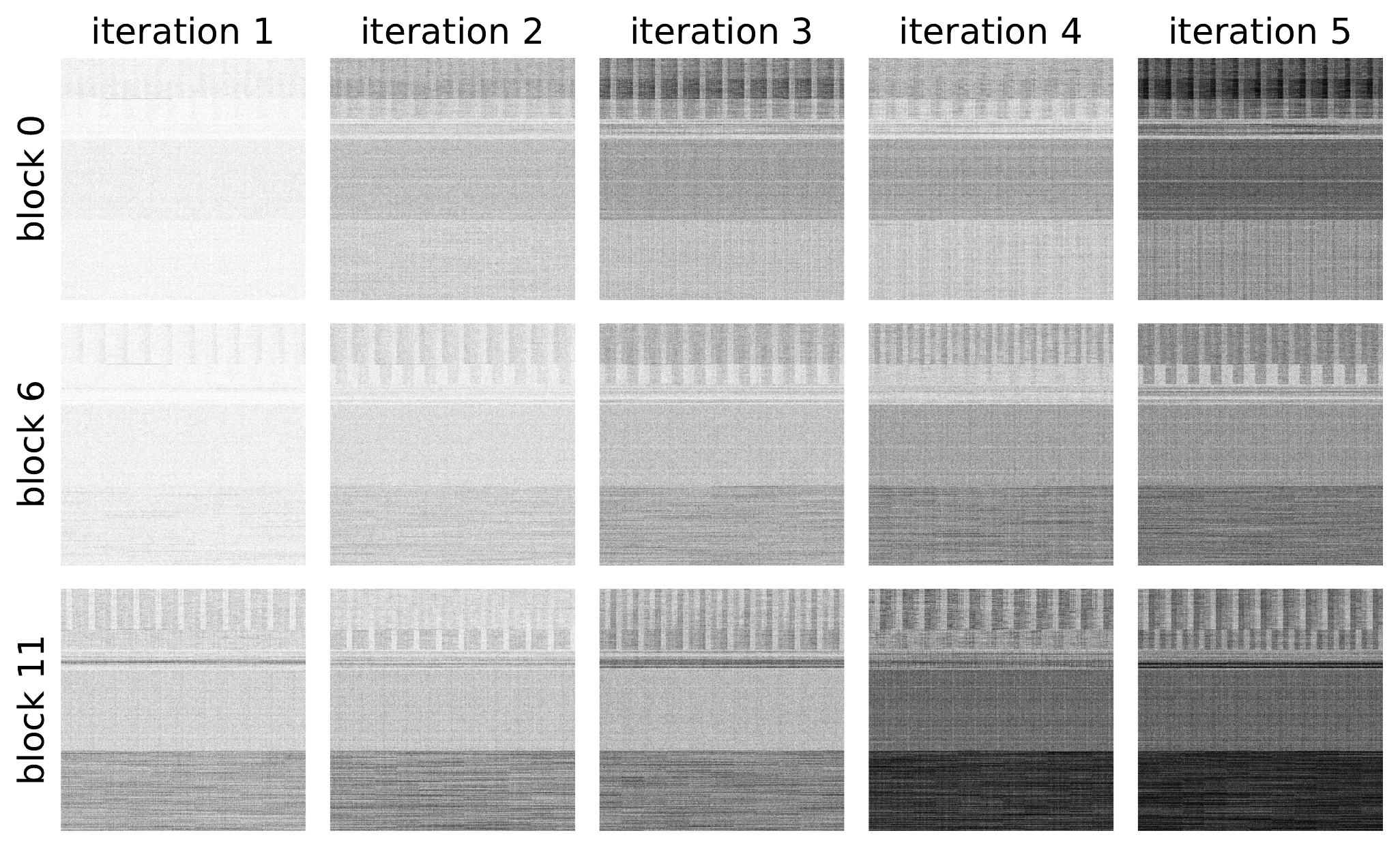}
	\caption{The visualization of gradient masks for blocks 0, 6, and 11 of the transformer encoder at different mask update iterations. Masks are reshaped into squares for visualization.}
	\label{fig:mask_evolution}
\end{figure}

In this section we conduct several qualitative studies to analyze the behavior of the proposed ViT-BERT model.

To demonstrate how images and text have a shared representation space in the transformer encoder, we illustrate in \autoref{fig:neuron_activation} the activation heatmaps of transformer blocks at different training steps. From \autoref{fig:neuron_activation}, we can see that before being jointly trained on images and text (iteration 0), all blocks in the transformer encoder have separate and scatter responses to images and text, implying that most neurons response to either images or text and are not shared across modalities. Along the joint training (iteration 500K and 1M), the neurons in the top layers of the transformer are gradually co-activated by both images and text, while bottom layers of the transformer still keep separate responses. This phenomenon is consist with our intuition that as the neural network go deeper, the features become more abstract and related to high-level semantics, thus less modality-specific. Therefore, neurons in the top layers might be activated by similar concepts from different modalities, \eg, a neuron can be activated by both a cat image and the word ``cat''.

To analyze how the iterative gradient masking strategy affects the proposed model during training, we first plot the sparsity of masks for all blocks in the transformer encoder at different mask update iterations, as shown in \autoref{fig:mask_sparsity}. We can see that the mask sparsity of the top blocks in the transformer quickly increases at the first iteration and they are overall higher than the mask sparsity of bottom blocks at the final iteration. We think this pattern of the mask sparsity is because the representation at top blocks is more abstract and compressed than that at bottom blocks, implying that the top blocks require fewer parameters to capture the information in the text pre-training and thus these blocks leave more parameters for the image pre-training. \autoref{fig:mask_evolution} shows how the gradient masks evolve during the training for the blocks 0, 6, and 11. It is clear that the mask of block 11 quickly becomes sparse (dark in \autoref{fig:mask_evolution}) at iteration 1 and it is much sparser than the mask of block 0 at iteration 5.

\section{Conclusions and Discussion}
In this paper we explore the direction of building a unified foundation model for multiple modalities. We take a step towards this goal by proposing a unified foundation transformer, \ie, ViT-BERT, which serves as a single pre-trained model for both the image and text domains. The proposed ViT-BERT model is intentionally designed to have minimal modality-specific and task-specific parameters and thus different domains can share as many parameters in the transformer encoder as possible. The ViT-BERT model is jointly pre-trained on unpaired images and text, and we also propose a knowledge distillation technique and an iterative gradient masking strategy to improve the performance of the joint pre-training. Extensive experimental results have demonstrated that the jointly pre-trained ViT-BERT model has learned high-quality representations for both the vision-only and text-only tasks. Ablation studies validate the importance of the proposed knowledge distillation and the gradient masking strategy. Qualitative studies also illustrate that the representation space at the top blocks of the transformer is highly shared across images and text.

\vspace{-10pt}
\paragraph{Limitations and Potential Negative Societal Impacts.} While we envision a unified foundation model that can serve various modalities, this work only focuses on images and text. There are more modalities to be considered, such as video and audio. Prior works (\eg, \cite{akbari2021vatt}) have shown the possibility of building a shared transformer for video, audio, and text, although it restrictively desires aligned data triplets from the three modalities and does not explore the rich text sources. Besides, we only consider single-modal tasks that are either vision-only or text-only. More techniques involving multi-modal pre-training (\eg, vision-language pre-training \cite{chen2020uniter}) or downstream multi-modal reasoning tasks (\eg, visual question answering or image captioning) might be incorporated to our work in the future.

Our model learns a shared representation space for the image and text modalities. Its predictions are based on the learned statistics from the training sets and will reflect biases from the used data, including ones that might have negative societal impacts like gender or racial biases.

{\small
\bibliographystyle{ieee_fullname}
\bibliography{egbib}
}

\clearpage
\appendix
\section{Experiment Details}
\subsection{Pre-training}
In the joint training baseline (``Joint''), we use a weighted sum of losses from the text and image pre-training tasks to train the model:

\begin{equation}
    \mathcal{L}_{\text{joint}} = \lambda \mathcal{L}_{\text{img}} + (1 - \lambda) \mathcal{L}_{\text{txt}}
\end{equation}
We perform a hyperparameter sweep to select the ratio $\lambda$ from $\{0.1, 0.2, 0.3, ..., 0.9\}$, as shown in \autoref{tab:lambda}. Based on the masked language modelling accuracy and the image classification accuracy of the validation sets in the pre-training, the best ratio $\lambda$ is selected as $0.2$.

\begin{table}[h]
\centering
\caption{Hyperparameter sweep of the ratio ($\lambda$) in the joint training baseline. ``NSP'' is short for ``Next Sentence Prediction'', ``MLM'' is for ``Masked Language Modelling'', and ``IC'' is for ``Image Classification''. ``Avg'' is the average of the MLM accuracy and the IC accuracy.}
\label{tab:lambda}
\begin{tabular}{c|cccc}
\shline
\textbf{Ratio ($\lambda$)} & \textbf{NSP} & \textbf{MLM} & \textbf{IC} & \textbf{Avg} \\
\hline
0.1            & 98.7                              & 64.4                               & 71.1                          & 67.8          \\
0.2            & 98.8                              & 63.6                               & 73.3                          & \textbf{68.5} \\
0.3            & 98.7                              & 63.4                               & 73.0                          & 68.2          \\
0.4            & 98.6                              & 62.9                               & 72.5                          & 67.7          \\
0.5            & 98.3                              & 61.7                               & 73.8                          & 67.8          \\
0.6            & 98.5                              & 60.5                               & 73.6                          & 67.1          \\
0.7            & 98.4                              & 60.2                               & 73.5                          & 66.9          \\
0.8            & 96.7                              & 54.6                               & 74.1                          & 64.4          \\
0.9            & 95.2                              & 46.2                               & 73.6                          & 59.9          \\
\shline
\end{tabular}
\end{table}

\subsection{Fine-tuning}
When fine-tuning the pre-trained models on a downstream task, we remove the original output heads and add a two layer, random-initialized MLP outputting the number of classes required by the target task.

For the fine-tuning on the vision-only tasks, we fine-tune all models using SGD with a momentum of 0.9 and run a grid search over the learning rates from $\{0.001, 0.003, 0.01, 0.03, 0.06\}$.

For the fine-tuning on the text-only tasks, we use the
AdamW optimizer and run a grid search over the learning rates from $\{5e^{-5}, 4e^{-5}, 3e^{-5}, 2e^{-5}\}$.

\section{Qualitative Examples}
\begin{figure*}[t]
	\centering
	\includegraphics[width=\linewidth]{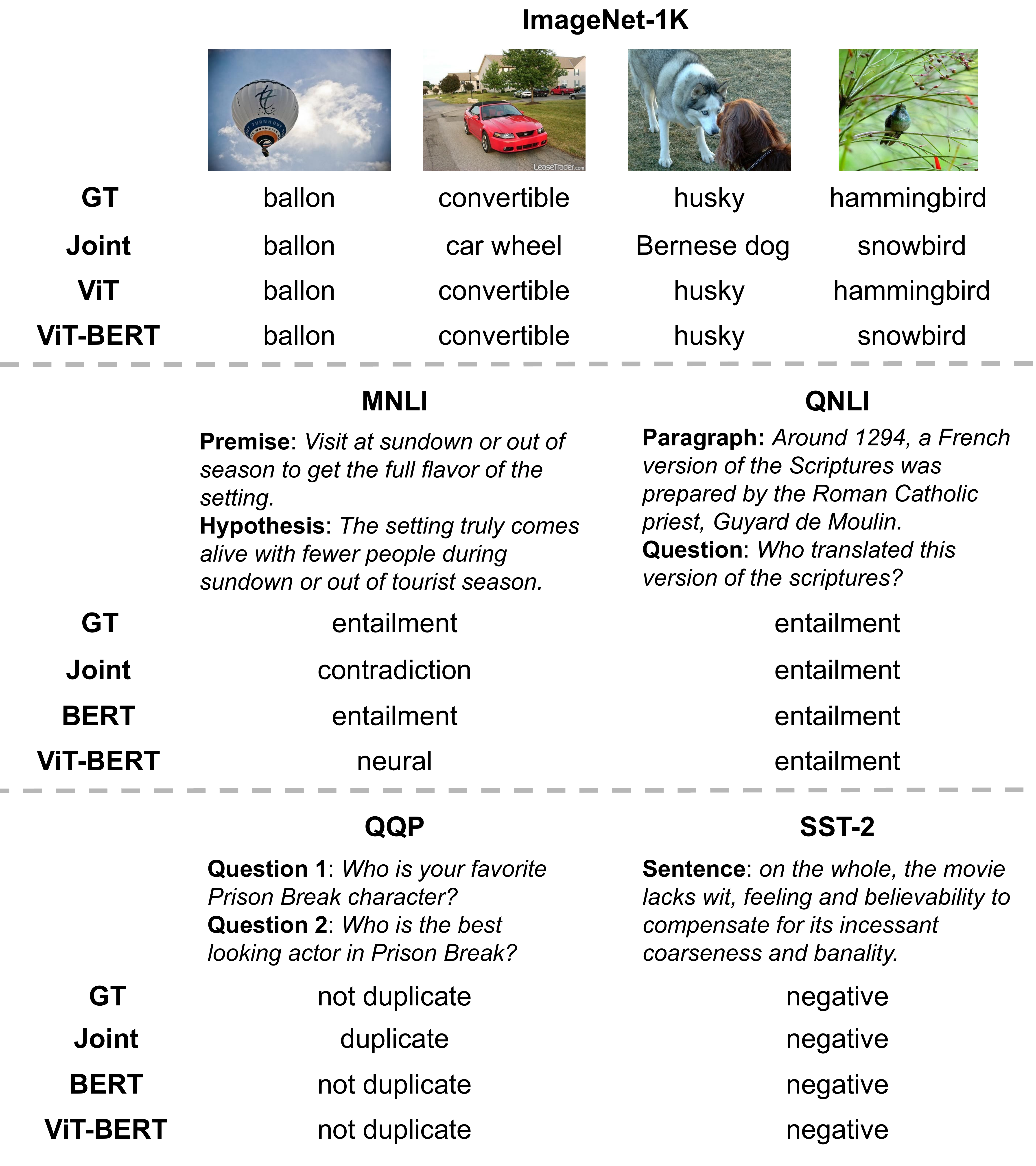}
	\caption{Predictions of our model and baselines across vision-only and text-only tasks. ``GT'' stands for the ground truth.}
	\label{fig:examples}
\end{figure*}

\autoref{fig:examples} shows qualitative examples of model predictions from the proposed ViT-BERT model and baselines across vision-only and text-only tasks.

\end{document}